\documentclass[sigconf]{aamas} 
\usepackage{makecell}
\usepackage{balance} % for balancing columns on the final page
\renewcommand\footnotetextcopyrightpermission[1]{}

\setcopyright{ifaamas}
\acmConference[AAMAS '23]{Proc.\@ of the 22nd International Conference
on Autonomous Agents and Multiagent Systems (AAMAS 2023)}{May 29 -- June 2, 2023}
{London, United Kingdom}{A.~Ricci, W.~Yeoh, N.~Agmon, B.~An (eds.)}
\copyrightyear{2023}
\acmYear{2023}
\acmDOI{}
\acmPrice{}
\acmISBN{}

\acmSubmissionID{636}

\title{
  Bringing Diversity to Autonomous Vehicles: An Interpretable Multi-vehicle Decision-making and Planning Framework
  }

\author{Licheng Wen, Pinlong Cai, Daocheng Fu, Song Mao and Yikang Li}
\affiliation{
  \institution{Shanghai AI Laboratory}
  \city{Shanghai}
  \country{China}}
\email{{wenlicheng,caipinlong,fudaocheng,maosong,liyikang}@pjlab.org.cn}

\begin{abstract}
  With the development of autonomous driving, it is becoming increasingly common for autonomous vehicles (AVs) and human-driven vehicles (HVs) to travel on the same roads. Existing single-vehicle planning algorithms on board struggle to handle sophisticated social interactions in the real world. Decisions made by these methods are difficult to understand for humans, raising the risk of crashes and making them unlikely to be applied in practice. Moreover, vehicle flows produced by open-source traffic simulators suffer from being overly conservative and lacking behavioral diversity. We propose a hierarchical multi-vehicle decision-making and planning framework with several advantages. The framework jointly makes decisions for all vehicles within the flow and reacts promptly to the dynamic environment through a high-frequency planning module. The decision module produces interpretable action sequences that can explicitly communicate self-intent to the surrounding HVs. We also present the cooperation factor and trajectory weight set, bringing diversity to autonomous vehicles in traffic at both the social and individual levels. The superiority of our proposed framework is validated through experiments with multiple scenarios, and the diverse behaviors in the generated vehicle trajectories are demonstrated through closed-loop simulations.
\end{abstract}


\newcommand{\BibTeX}{\rm B\kern-.05em{\sc i\kern-.025em b}\kern-.08em\TeX}

\begin{document}

%%% The following commands remove the headers in your paper. For final 
%%% papers, these will be inserted during the pagination process.

\pagestyle{fancy}
\fancyhead{}

%%% The next command prints the information defined in the preamble.

\maketitle

%%%%%%%%%%%%%%%%%%%%%%%%%%%%%%%%%%%%%%%%%%%%%%%%%%%%%%%%%%%%%%%%%%%%%%%%

\section{Introduction}
%FIXME: introduction中文献说的过于具体了，可能还是单开一个related works好一些？

With the combined efforts of academia and industry, research on autonomous driving has flourished over the past decade.
Nowadays, a growing number of companies are testing their autonomous vehicles (AVs) on the road, becoming new participants in the traffic flow besides human-driven vehicles (HVs).
The real-world vehicle flow encompasses a diversity of behaviors at both the social and individual levels, known as \textit{social behavior} and \textit{driving habit}.
Social behavior~\cite{Wang2022,Schwarting2019} implies how a vehicle interacts with others. When another vehicle changes lanes, overtakes or merges, the driver chooses whether to persist in the current movement or yield according to experience.
Drivers also have their own characteristics when driving. Therefore driving habits~\cite{ranney1994models,wang2018influence,wang2022modeling} is introduced to describe the individual difference, especially on cruising comfort and driving safety.

\begin{figure*}[htbp]
  \centering
  \includegraphics[width=\linewidth]{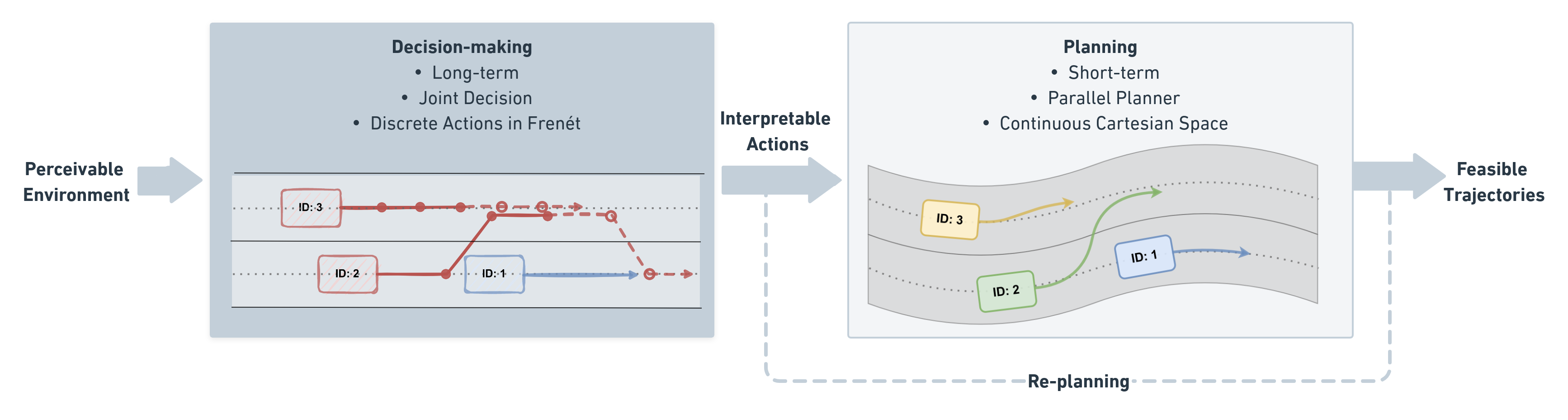}
  \caption{Schematic diagram of the proposed decision-making and planning framework}
  \label{fig:framework}
  \Description{A framework of the proposed decision-making and planning structure.}
  \vspace{10pt}
\end{figure*}

The behavioral diversity of AVs is mainly expressed through the decision-making and planning modules, which are core components of the automated driving system.
The prevailing single-vehicle decision-making and planning approaches can be divided into two categories: heavy-decision-based frameworks and light-decision-based ones~\cite{KATRAKAZAS2015416, claussmann2019review}. The former methods typically separate decision-making and motion planning to reduce the computational burden and complexity~\cite{urmson2008autonomous, montemerlo2008junior}, whereas the light-decision-based ones improve the algorithm's adaptability in corner cases by weakening the decision-making module and increasing the planner sampling size~\cite{Howard2007,Fan2018b,Ding2019c}.
However, these approaches struggle to game with other traffic participants in the real world and therefore exhibit an insufficient understanding of social interactions.
% Mutually, this also makes it challenging for human drivers to understand the behaviors of AVs.
% lack Interactive

As vehicle-to-everything (V2X) technologies take hold in self-driving vehicles, centralized decision-making and trajectory planning for multiple vehicles in traffic become possible.
The centralized planner addresses the complex interactions in the vehicle flow effectively, including competition and collaboration.
There are several types of methods for multi-vehicle decision and planning.
Optimization-based methods~\cite{sadigh2016planning,fisac2019hierarchical,hang2022driving} provides good modelling of game-theoretic interactions between vehicles.
Methods based on the Monte Carlo tree search~\cite{Browne2012,Lenz2016,Li2022} employ a tree structure to manage the interactions and considerably improve the algorithm's efficiency by sampling vehicles' action space. However, these methods output the long-term control variables directly, which can be risky when there are  HVs on the road.
Learning-based approaches~\cite{Casas2021,Peng2021a,Chen2022,sun2022m2i} obtain vehicles' driving strategies from open datasets and avoid collisions by predicting other cars' trajectories.
The trajectories generated by such methods fail to express driving intentions explicitly and are therefore poorly understood by passengers or other road users. They also rely heavily on datasets.
% lack Interpretable

In addition, popular open-source autonomous driving simulators provide solutions for background traffic generation.
SUMO~\cite{SUMO2018} applies a car-following model and a lane-changing model to each vehicle. The trajectories produced by this method can avoid collisions but lack vehicle kinematic constraints.
CARLA~\cite{Carla2017} adopts a traffic manager module to generate commands for vehicles according to the current simulation state.
However, this approach applies rigid behavioral models, which leads to excessively conservative and non-diverse trajectories.
% lack Diversity

We believe that modeling behavioral diversity provides a better understanding of real-world traffic flows and leads to a more credible autonomous-driving simulation.
We propose a hierarchical multi-vehicle decision and planning framework, where the high level makes joint decisions for all vehicles in the scenario, and the low level generates kinematically trajectories for each controlled AV.
% The high level of the framework presents a modified version of the Monte Carlo tree search to make joint behavioral decisions for all vehicles in the scenario, while in the low level, we parallel the planning module to generate kinematically feasible trajectories for each controlled AV. 
The key features of our framework can be summarized as \textbf{DII}:
\begin{itemize}
  \item \textbf{D}iversity: By adding a vehicle cooperation factor in the decision module and introducing the trajectory weight set to the planning module, our framework produces solutions with social and individual-level diversity.
  \item \textbf{I}nterpretability: The decision module in our framework produces interpretable action sequences for each AV to keep the passenger on board aware of the current situation and explicitly communicate intentions to the surrounding HVs through brake lights and turn signals.
  \item \textbf{I}nteractivity: We include all vehicles in the traffic flow when making decisions and can therefore explore complex interactions between them. Moreover, the framework ensures a real-time response to the dynamic environment through high-frequency replanning.
\end{itemize}

% 可以不要
The remainder of this paper is organized as follows. Section \ref{sec:framework} describes the framework of our proposed methodology.
Section \ref{sec:methodology} introduces the details of the framework, including multi-vehicle Monte Carlo tree search considering social behavior and parallel trajectory planner incorporating driving habits.
Section \ref{sec:experiment} presents the performance test along with closed-loop simulations and discusses the results.
Finally, the conclusion and future work are presented in Section \ref{sec:conclusion}.

\section{Framework}
\label{sec:framework}

Consider a set $\mathcal{V} = \{V_1,V_2,\cdots,V_N \}$ containing $N$ vehicles including AVs and HVs. There are $K$ AVs available for centralized planning, denoted as $\mathcal{V_{C}} = \{V_1,\cdots,V_k \}\subseteq \mathcal{V}$.
The remaining are AVs and all HVs are considered as uncontrolled vehicles that make decisions and motion planning independently.

Our task is to generate feasible trajectories with behavioral diversity for each controlled vehicle $V_i \in \mathcal{V_{C}}$ and collaborating with uncontrolled vehicles $V_j \in \{\mathcal{V} \setminus \mathcal{V_{C}}\}$ in the environment. We introduce a two-stage multi-vehicle framework containing both decision-making and trajectory planning modules.
The input to the framework is a perceivable environment which contains road conditions, route information, vehicle status and predictions of uncontrolled vehicles.
The schematic diagram of our proposed framework is illustrated in Figure \ref{fig:framework}, where vehicles with IDs 2 and 3 are AVs controlled by our framework while vehicle 1 is an uncontrolled vehicle.
The details of the two stages in our proposed framework are described below.

In the first stage, we propose a decision module to address the social interactions among vehicles. This module generates long-term, coarse decisions for all vehicles jointly and simultaneously.
The environment is constructed on the Frenét frame~\cite{Rovenski2006} for decision-making, which allows the decision module to focus on inter-vehicle interactions without considering road constraints.
Since actions like continuous lane-changing and overtaking take a relatively long time to complete, the decision module is required to be forward-looking and consistent.
The decision module solves the interaction game between vehicles by making centralized joint decisions on the vehicle flow. It can generate diverse social behaviors in the flow via setting different cooperation factors in the reward function.
The decision module generates a temporal sequence of discrete actions for each controlled vehicle. Since each action in the sequence has a physical meaning, such as changing lanes, accelerating, maintaining speed, etc., the output of our decision module is interpretable.

In the second stage, the planning module receives discrete action sequences from the first stage and generates continuous kinematic-feasible trajectories for each controlled vehicle.
Thanks to the forward-looking nature of our front-end decision module, the trajectory planning module only needs to generate trajectories within a short planning horizon.
Our planning module adopts a distributed parallel architecture, i.e. running a planner independently for each vehicle, which more closely resembles human driving in the real world.
For each parallel planner, the leading part of its received action sequence is selected as the guidance to generate a continuous trajectory that conforms to the vehicle kinematic constraints in Cartesian space.
This module reflects the driving habits of different drivers through planner's selection strategy for various trajectories.
It is worth emphasizing that the planner would maintain a high replanning frequency and continuously predicts surrounding vehicles to avoid potential collisions.

% 我们的模块是具有通用性的，可以替换
Our proposed two-stage framework provides an interpreted result with physical meaning for each module,allowing the AV to inform passengers of the current situation and to explicitly communicate its own intentions to the HV surrounding them.
Meanwhile, it guarantees bottom-line security. The planning module realizes collision risks by its prediction of other vehicles and generates collision-free trajectories, even if the decision module produces some impractical action sequences.

\section{Methodology}
\label{sec:methodology}
\subsection{Joint Multi-vehicle Decision-making}

\subsubsection{Basic Monte-Carlo Tree Search Algorithm}
\begin{figure}[tbp]
  \centering
  \includegraphics[width=\linewidth]{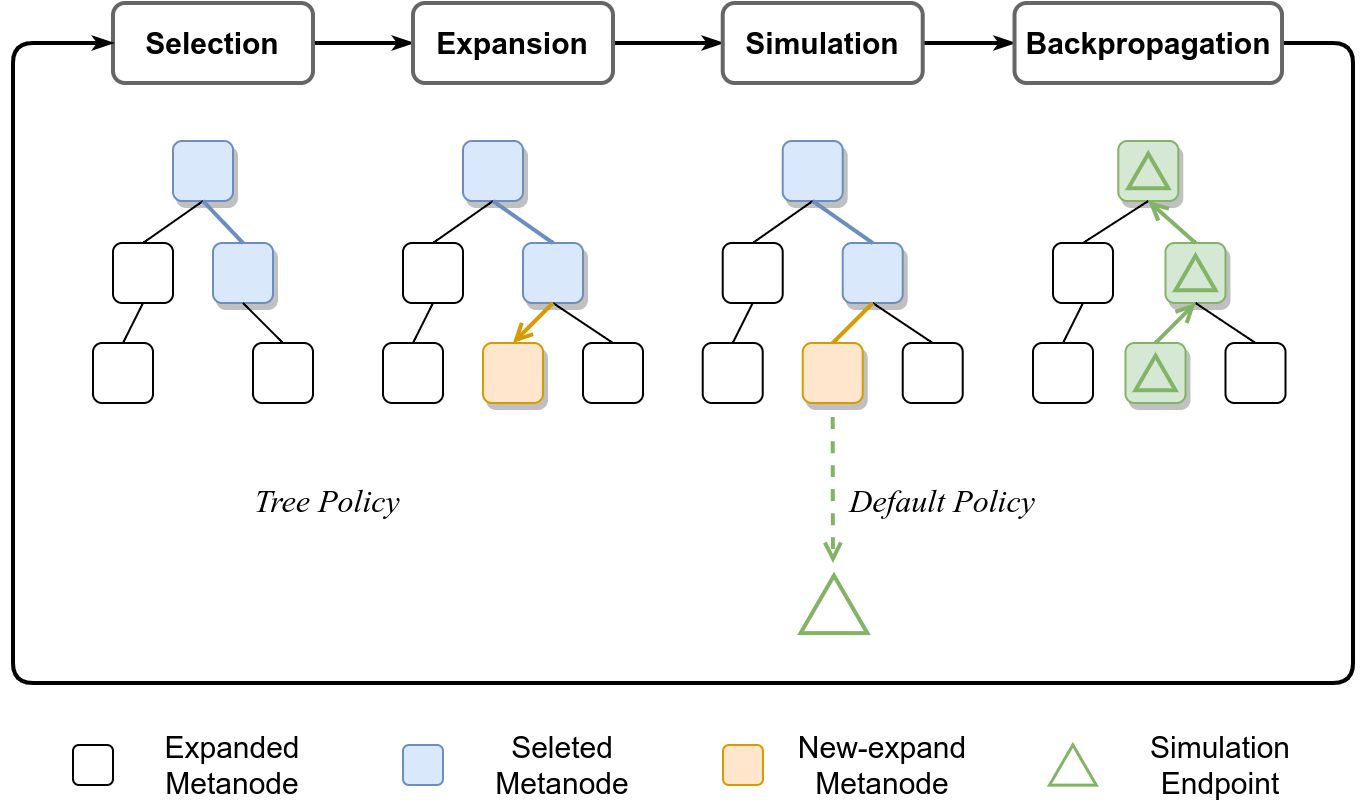}
  \caption{Procedure of the Monte-Carlo Tree Search}
  \label{fig:mcts general}
  \Description{A demo of the structure of the MCTS algorithm.}
\end{figure}

Monte-Carlo Tree Search, also known as MCTS, is a heuristic search algorithm for the decision process which analyses the search space for the most promising actions based on the Monte Carlo sampling method.
The general MCTS algorithm~\cite{Browne2012} includes four phases in one search iteration: selection, expansion, simulation, and back-propagation, as shown in Figure~\ref{fig:mcts general}.

The selection phase is performed based on an existing search tree. The algorithm will recursively select the most valuable child node using a tree policy until reaching a non-fully expanded node.
Then, the selected node is expanded with one of the remaining possible actions to get a new leaf node.
From this node, the algorithm will use a default policy (also called roll-out) to perform the simulation until it reaches a terminal node.
At the back-propagated phase, the algorithm will extract the rewards from the simulation sample and then use them to update the average rewards through all nodes that have been traversed.
After reaching predefined maximum iterations, the solution can be obtained by greedily selecting the child node with a maximum reward at each layer.

The standard UCT algorithm~\cite{Browne2012} is applied in order to address the exploration-exploitation dilemma in the selection phase. The UCT value of child node $j$ can be calculated by:
\begin{eqnarray}
  \label{eq:UCT}
  UCT&=&\overline{X}_j+2C_p\sqrt{\frac{2\ln n}{n_j}}
\end{eqnarray}
where $\overline{X}_j$ is the average reward of node $j$, $n$ is the number of times the parent node of $j$ has been visited, $n_j$ is the number of times child j has been visited, and $C_p>0$ is a constant to balance exploration and exploitation. The reward distribution of each node should lie between 0 and 1.

The vanilla MCTS algorithm typically makes decisions for a single agent. However in our framework, the decisions are required for all vehicles in the flow at the same time.
A trivial approach would be to perform a separate MCTS search for each vehicle. However performing MCTS searches separately for each vehicle requires accurate predictions of other vehicles' behavior before making a decision,. This violates Vapnik's Principle~\cite{Vapnik2006} that"When solving a problem, don't solve a harder problem as an intermediate step.", as modeling and predicting the behavior of other agents is a harder problem.

\subsubsection{Metanode in the MCTS}
Several multi-vehicle decision methods based on the Monte Carlo tree search have been developed.
Lenz et al.~\cite{Lenz2016} propose a method for generating anticipatory and cooperative behavior of AVs, but its direct output of motion via MCTS is poorly robust to dynamic environments, especially with HVs mixing in the road.
Another approach~\cite{Li2022} sets priorities for the vehicles and makes decisions from highest priority to lowest, with the subsequent agent required to satisfy the decisions completed by the preceding ones. However, such approaches fail to express the different social behaviors in the flow, as the low-priority vehicles only passively respond to the decisions made by the high-priority ones. Besides, it's never easy to assign priority to every single vehicle in the traffic flow.

\begin{figure}[tbp]
  \centering
  \includegraphics[width=0.95\linewidth]{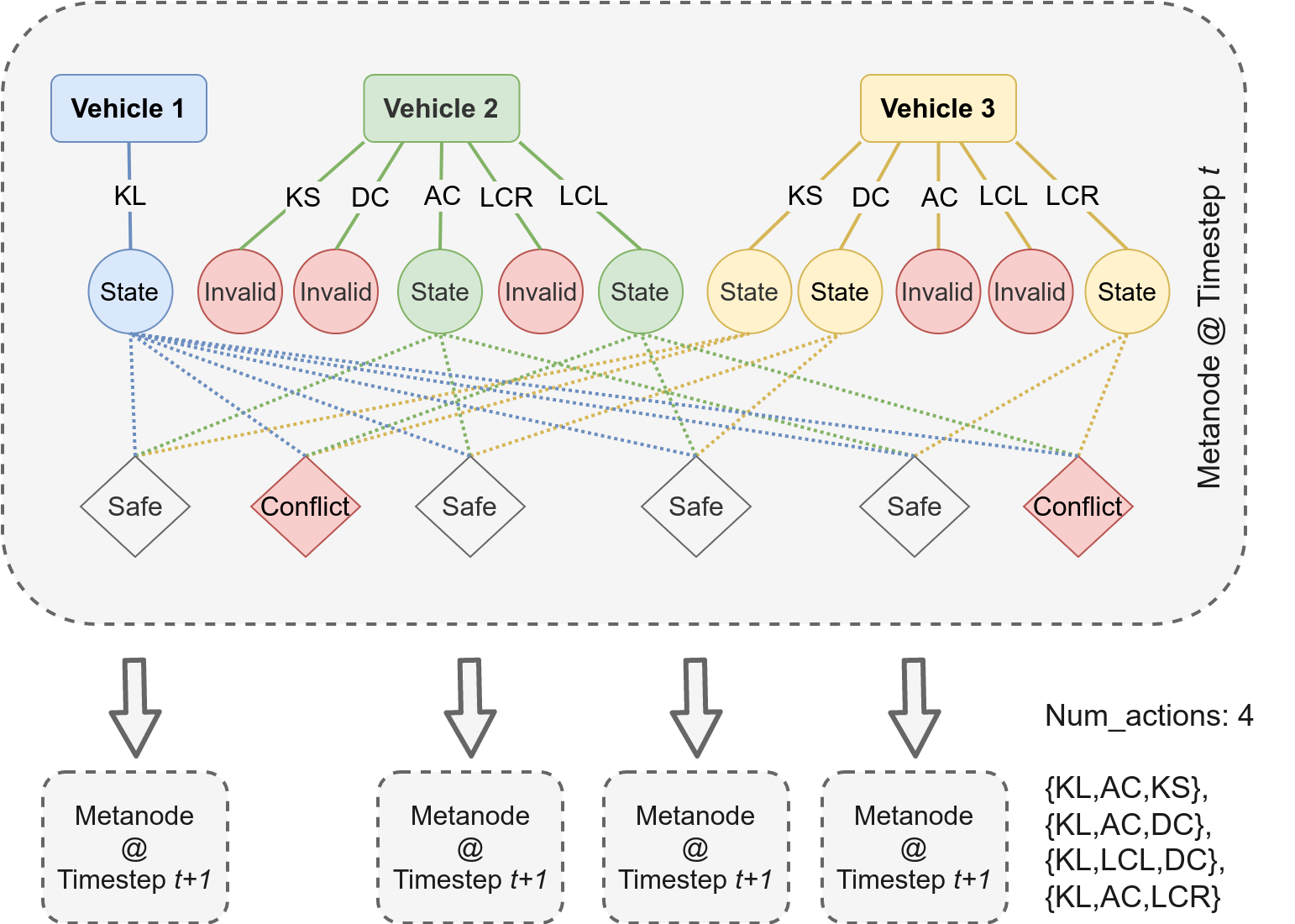}
  \caption{A metanode at time step $t$ in the MCTS}
  \label{fig:mcts metanode}
  \Description{A metanode at time step $t$ in the MCTS search.}
\end{figure}

Instead, our approach takes the HVs in the traffic flow into account and there is no priority between vehicles in the decision-making process.
We replace the node in the general MCTS method with a \textit{metanode} that can generate multi-vehicle actions simultaneously. The metanode's extension is shown in Figure \ref{fig:mcts metanode}, where vehicle 2 and 3 are vehicles controlled by our framework while vehicle 1 is an uncontrolled vehicle.

For a metanode extended to time step $t$, it receives all controlled vehicle's state at the current moment. In particular, the state for vehicle $V_i$ at time step $t$ can be described as $\mathbf{x}_i^t = [s,d,v]^T$, $s$ and $d$ are the Frenét coordinates of the vehicle and $v$ is the longitudinal velocity.
Similar to \cite{Lenz2016}, we define a set of possible actions for each controlled vehicle to take in one time step. As listed in Table \ref{tab:actions}, there are five actions - maintain speed (KS), accelerate (AC), decelerate (DC), change lanes to the left (LCL) and to the right (LCR). By selecting different actions, the vehicle can then calculate its corresponding state at the next time step.
As for uncontrolled vehicles in the flow, we assume they performing a lane-keeping (KL) action.  This action dynamically adjusts their velocity and always maintains an appropriate distance from the leading vehicle during the decision-making process.

\begin{table}[tbp]
  \caption{Actions available for vehicles in one time step}
  \label{tab:actions}
  \resizebox{0.95\linewidth}{!}{
    \begin{tabular}{cll}\toprule
      Action & State change                                                            & Description                                   \\ \midrule
      KS     & $\left[v\Delta t,0,0\right]$                                            & \makecell[l]{Maintain the current velocity}   \\
      AC     & $\left[v\Delta t+\frac{a_{acc}}{2}\Delta t^2,0,a_{acc}\Delta t\right]$  & \makecell[l]{Increase the current velocity    \\ with a fixed acceleration $a_{acc}$} \\
      DC     & $\left[v\Delta t-\frac{a_{dec}}{2}\Delta t^2,0,-a_{dec}\Delta t\right]$ & \makecell[l]{Decrease the current velocity    \\ with a fixed deceleration $a_{dec}$} \\
      LCL    & $\left[v\Delta t,-\Delta d,0\right]$                                    & \makecell[l]{Make a partial left lane change  \\  with a width of $\Delta d$} \\
      LCR    & $\left[v\Delta t,\Delta d,0\right]$                                     & \makecell[l]{Make a partial right lane change \\  with a width of $\Delta d$}  \\
      KL     & -                                                                       & \makecell[l]{Keep the current lane            \\(Only for uncontrolled vehicles)} \\
      \bottomrule
    \end{tabular}
  }
\end{table}

Since decisions need to be made for all vehicles simultaneously within a metanode, the output of the metanode would combine each vehicle's possible behavior.
As a result, the search tree grows exponentially when the MCTS gradually evaluates the whole interaction decision space. Finding the best solution that is computationally feasible becomes impossible in practice. Therefore pruning the expansion of the metanode proves necessary.

\subsubsection{Pruning in MCTS}
\label{prune strategy}
Although each vehicle has five optional actions, not every one of them is feasible and some may lead to a potential collision.
Obviously, vehicles on either side of the road cannot continue to make lane-changing actions to over borderlines.
According to \cite{Swaroop2001}, vehicles in traffic should maintain a shortest safe distance $D_{s}$ from its leading vehicle. $D_{s}$ consisted of two parts: reaction distance and minimum headway, which are defined as:
\begin{eqnarray}
  \label{eq:safe distance}
  D_{s}&=& v \cdot \tau + MTH\cdot \Delta v
\end{eqnarray}
where $\tau>0$ and $MTH>0$ are both time constants, representing reaction time and minimum time headway respectively. $v$ denotes vehicle longitude velocity and $\Delta v= v-v_l$ denotes the velocity difference between the self and lead vehicles.
Further, the velocity limit that any single vehicle in the flow should satisfy is:
\begin{eqnarray}
  \label{eq:vel limit}
  v&\in&\left[v_{f}-\frac{\Delta s_{f}-\tau\cdot v_f}{{MTH}},\min(\frac{{MTH}\cdot v_{l} +\Delta s_{l}}{\tau + {MTH}},\frac{\Delta s_{l}}{\tau})\right]
\end{eqnarray}
where $\Delta s_{l}$ and $\Delta s_{f}$ indicate the gap between the current vehicle and its leading/following vehicle, respectively. $v_{l}$ and $v_{f}$ denote the velocity of the leading and following vehicle.
It is worth noting that for the upper limit, even if the vehicle's current velocity is slower than the leading vehicle, it still needs to maintain a minimum reaction distance.

In the metanode at time step $t$, when vehicle $i$ takes an action $a$ from Table~\ref{tab:actions} and the resulting state $\mathbf{x}_i^{t+1}$ does not satisfy Equation~(\ref{eq:vel limit}), the action $a$ becomes invalid and the corresponding state $\mathbf{x}_i^{t+1}$ will be abandoned.
Also for all valid state combinations within the metanode, the algorithm checks for collisions between new states in each combination and for conflicts in the process of reaching the states (e.g. two vehicles changing into each other's lanes). If a conflict occurs, the action combination should not be extended for the metanode at time step $t+1$.
Through these techniques, the decision space is significantly reduced and the search efficiency is guaranteed. As Figure~\ref{fig:mcts metanode} shows, these three vehicles originally generated $5^2=25$ possible action combinations, and only 4 valid combinations remain after pruning. The efficiency of the algorithm improves more significantly when facing complex traffic situations.

\subsubsection{Reward Function with Social Behavior}
During the simulation phase of MCTS, the algorithm continues to construct the metanode, but only one valid action combination is selected to extend the next time step until the simulation terminates.
Each simulation can be terminated either when all vehicles have traveled a certain distance in their target lane, or when the simulation reaches the maximum time step with any vehicles not completing their intentions.
Then a reward for this simulation should be calculated from the result.

We first calculate the reward $R_i$ separately for each vehicle $i$ in the traffic flow.
The reward $R_i$ has three parts, namely driving in the target lane, driving in the lane's center line and maintaining the consistency of the actions. Besides, the reward should meet in the range of 0 to 1.

For the vehicle flow to exhibit a diversity of social behaviors, the cooperative tendency of the vehicle needs to be characterized by the reward.
Similar to \cite{Schwarting2019}, we introduce a behavior reward considering both reward to self and reward to others:
\begin{eqnarray}
  R &=& R_{\rm{self}}~+~\gamma~R_{\rm{other}}
\end{eqnarray}
where $\gamma\in[0,1]$ being the cooperation factor.
$\gamma=0$ implies the vehicle is egoistic and takes no account of the behavior of other vehicles, whereas $\gamma=1$ denotes the vehicle treats other vehicles equally important to itself when making decisions. Each vehicle $i$ in the flow possesses its unique factor $\gamma_i$.

Since each metanode in the search tree handles the actions of the whole vehicle flow,  the reward of the simulation is obtained by combining  all vehicles' behavior rewards:
\begin{eqnarray}
  X_{\rm{flow}}&=&\frac{1}{K}\sum_{1\leq i \leq K} \frac{R_i+\gamma_i \sum_{j\neq i}{R_j}}{1+(K-1)\gamma_i}
\end{eqnarray}
Since $R_i \in [0,1]$, the distribution of $X_{\rm{flow}}$ is also guaranteed to stay in $[0,1]$.
Finally, reward $X_{\rm{flow}}$ is used to update the average rewards of all selected metanodes in the back-propagation phase.

\subsection{Parallel Vehicle Trajectory Planning}

The decision module gives a coarse-grained action sequence for each vehicle, but these actions take no account of vehicle kinematics and road curvature. They thus cannot be performed by the vehicles directly.
Meanwhile, the generated discrete actions may still expose the vehicles to potential collision risk.
We employ a trajectory planning module after decision-making to generate continuous, kinematic feasible trajectories and ensure the bottom-line safe of the whole framework.

\subsubsection{Conversion between Cartesian and Frenét Frame}
As mentioned earlier, we perform decision-making in the Frenét coordinate system to simplify the calculations.
However, trajectories generated by the planning module require converting to the global Cartesian frame for verifying the vehicle's kinematic constraints.
Here we briefly describe the conversion from the Frenét frame to the Cartesian coordinates.

% \begin{figure}[tbp]
%   \centering
%   \includegraphics[width=0.9\linewidth]{figures/frenet.png}
%   \caption{Trajectory in a Frenét Frame~\cite{Werling2010c}}
%   \label{fig:Frenét frame}
%   \Description{A Trajectory in a Frenét Frame.}
% \end{figure}

A Frenét coordinate system is constructed via a reference path generated by the center-line of the current lane, it is denoted by the arc-length parameter $s$ along the reference path and the lateral offset $d$. $\dot{s}$ and $\dot{d}$ are the derivative with respect to time.
%  as shown in Figure~\ref{fig:Frenét frame}. 
For a vehicle with the Frenét coordinates $\left[s,\dot{s},d,\dot{d}\right]$, we can calculate its position $(x,y)$, velocity $v$, orientation $\theta$ in the Cartesian frame according to the transformation proposed in~\cite{Werling2010c}:
\begin{eqnarray}
  \label{eq:Frenét}
  \left\{\begin{array}{lc}
    x=      & x_{r}-d \sin \theta_{r}                                                \\
    y=      & y_{r}+d \cos \theta_{r}                                                \\
    v=      & \sqrt{\left[\left(1-\kappa_{r} d\right)\dot{s}\right]^{2}+\dot{d}^{2}} \\
    \theta= & \arcsin \left(\frac{\dot{d}}{v}\right)+\theta_{r}                      \\
  \end{array}\right.
\end{eqnarray}
where $(x_r,y_r)$, $\theta_r$ and $\kappa_r$ denote the position, the orientation and curvature of the reference path at length $s$.

\subsubsection{Feasible Trajectory Planner}
\begin{figure}[tbp]
  \centering
  \includegraphics[width=0.9\linewidth]{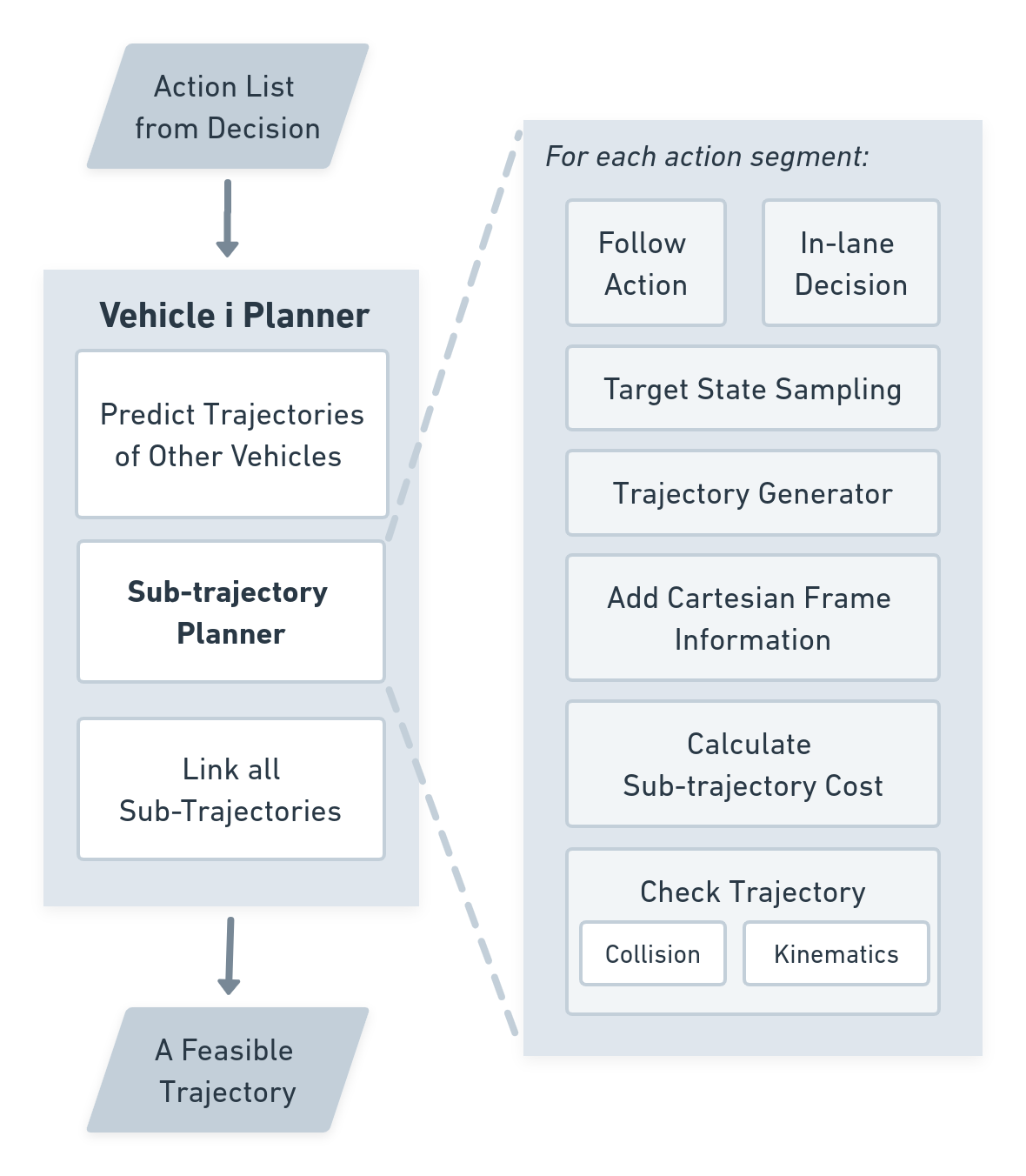}
  \caption{The pipeline of a single-vehicle trajectory planner}
  \label{fig:trajectory planner}
  \Description{This diagram demonstrates the pipeline of a vehicle trajectory planner.}
\end{figure}

As mentioned in Section~\ref{sec:framework}, in the planning module we employ a parallel architecture for trajectory planning of each vehicle, which not only improves the runtime efficiency of the framework but also resembles the way humans drive in the real world.
The pipeline of the single-vehicle trajectory planner is shown in Figure~\ref{fig:trajectory planner}.

For an AV $V_i \in \mathcal{V_{C}}$ under our control, it receives an action sequence from the decision module. Then one of the parallel planners takes over.
The planner first predicts the future trajectories of other vehicles in the flow.
Then it splits the different actions in the sequence and plans a sub-trajectory for each action segment.
Finally, the planner connects all the sub-trajectories to obtain the final feasible and continuous vehicle trajectory.

Our sub-trajectory planner is similar to the one proposed in \cite{Werling2010c}, but with some differences in the pipeline.
First, the algorithm determines the range of sub-trajectory's target state based on the current action segment. Although the state change for each action is a constant when making decisions as shown in Table~\ref{tab:actions}, we extend it to a continuous region of target states for sub-trajectory planning. Additionally, for vehicles with lane-keeping actions, the planner creates a larger target state region to make rational in-lane decisions.
The planner then uniformly samples a number of target states within the predefined region.
We employ quintic polynomial curves in the Frenét frame to generate the jerk-optimal connections between the start state and all sampled target states of the sub-trajectory.
After that, we convert each sub-trajectory to Cartesian coordinate by using Equation~\ref{eq:Frenét}, including its position, velocity and heading angle.
All alternative sub-trajectories are calculated with a cost function, which is detailed in Section~\ref{sec:cost function}.
We check whether each sub-trajectory meets the kinematic constraints and is collision-free with other vehicles in the flow.
The sub-trajectory that satisfies all the constraints with the smallest cost is selected and the final state of this sub-trajectory is set as the initial state for the next sub-trajectory.

\subsubsection{Cost Function considering Driving Habits}
\label{sec:cost function}

For each alternative sub-trajectory, a cost function should be  calculated to evaluate its safety and comfort.
In particular, there are several cost terms in the function:
\begin{itemize}
  \item Curve energy cost: $J_{i,cur} = \sum \kappa_i^2$, $\kappa_i$ denotes the curvature of a trajectory point. This penalty trajectory with sharp turns.
  \item Heading difference cost: $J_{i,\phi} = \sum \left|\phi_i  - \phi_r \right|^2$,$\phi_i$ denotes the orientation angle of the vehicle and $\phi_r$ indicates the road's direction at the point.
  \item Out of line cost: $J_{i,out} = \sum \left| d_i \right|^2$, $d_i$ denotes the lateral offset in Frenét frame. This term is only active for actions except for lane-changing ones.
        % \item Velocity deviation cost: $J_{i,vel} = \sum \left|v_i  - v_0 \right|^2 $
  \item Acceleration cost: $J_{i,acc} = \sum  a_i^2$, $a_i$ denotes vehicle's acceleration at a trajectory point.
  \item Jerk cost: $J_{i,jerk} = \sum  j_i^2$, $j_i$ denotes vehicle's jerk (derivative of acceleration) at a trajectory point. This term and the acceleration cost both contribute to the comfort of the trajectory.
  \item Dynamic obstacle cost: $J_{i,obs}$. This term serves to avoid the vehicle from being too close to the surrounding vehicles, which is described in detail below.
\end{itemize}

\begin{figure}[tbp]
  \centering
  \includegraphics[width=0.93\linewidth]{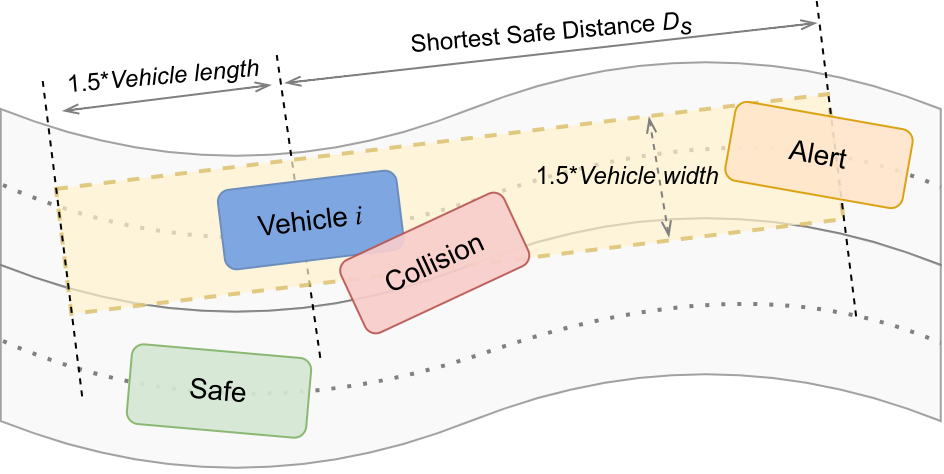}
  \caption{The alert zone of vehicle $V_i$}
  \label{fig:collison cost}
  \Description{This diagram demonstrates three different scenarios when calculating dynamic obstacle cost.}
\end{figure}

Dynamic obstacle cost takes the distance between a vehicle following a trajectory and dynamic obstacles on the road (i.e., surrounding vehicles)  into account. The cost for vehicle $V_i$ is defined as:
\begin{eqnarray}
  J_{i,obs} &=& \sum_{t=0}^{T}\sum_{\forall j\neq i}J_{i,obs}(t,j)
\end{eqnarray}
where $J_{i,obs}(t,j)$ characterizes the position of vehicle $V_j$ with respect to the current vehicle $V_i$ at time step $t$. As shown in Figure~\ref{fig:collison cost}, we introduce an alert zone whose shape is a rectangle surrounding the body of the vehicle $V_i$.
The zone has a length of the shortest safe distance $D_s$ defined by Equation~\ref{eq:safe distance} in front of the vehicle and a length of 1.5 times the body length behind the vehicle. The width of the alert zone is 1.5 times the width of the body.

The position of the vehicle $V_j$ in the flow relative to $V_i$ can be divided into three different types:  outside the alert zone, inside the alert zone without collisions, and having collisions with the current position of $V_i$.
The vehicle $V_j$ is safe outside the alert zone and therefore cost-free, whereas its direct collision with the trajectory is unacceptable. In the alert zone, the closer vehicle $V_j$ is to $V_i$, the higher the cost.
In particular, $J_{i,obs}(t,j)$ is defined as follows:
\begin{equation}
  J_{i,obs}(t,j)~=~
  \begin{cases}
    0,                                                & V_j~\rm{outside~alert~zone} \\
    C_z(2-\frac{|D|_x}{D_s+1.5l}-\frac{|D|_y}{1.5d}), & V_j~\rm{in~alert~zone}      \\
    \infty,                                           & V_j~\rm{has~collision}
  \end{cases}
\end{equation}
where $|D|_x$ and $|D|_y$ represent the component of the shortest distance from the vehicle $V_j$ to vehicle $V_i$ along the length and width of the alert zone, respectively. $l$ and $d$ denote the length and width of vehicle body and $C_z>0$ is a cost constant.

All the aforesaid terms are summed up as the cost function:
\begin{eqnarray}
  J_i  &=& \sum~w_{(\cdot)}~J_{i,(\cdot)}
  % \\
  % J_i &=& W_i \cdot [J_{i,cur},~J_{i,\phi},~J_{i,out},~J_{i,acc},~J_{i,jerk},~J_{i,obs}]^{\mathrm{T}}
\end{eqnarray}
where $J_{i,(\cdot)} \in \{J_{i,cur},~J_{i,\phi},~J_{i,out},~J_{i,acc},~J_{i,jerk},~J_{i,obs}\}$ represents all the cost items mentioned above, and $w_{(\cdot)}$ is the weight parameter corresponding to each cost item.
We observe different drivers in traffic flow may choose various trajectories depending on their driving habits. In other words, drivers weigh up these trajectory's cost terms differently. For instance, some drivers may be more concerned with driving comfort, while others prefer to keep a longer safe distance from other vehicles.
Therefore, we define a trajectory weight set $\mathcal{W}$ containing $M$ weight vectors based on the driver's driving habits:
\begin{eqnarray}
  \mathcal{W} = \{\mathbf{w}_1,\mathbf{w}_2,\cdots\mathbf{w}_M \}
\end{eqnarray}
Each weight vector $\mathbf{w}$ contains a unique combination of weights for all cost terms, $\mathbf{w} = [w_{cur},~w_{\phi},~w_{out},~w_{acc},~w_{jerk},~w_{obs}]$.
AVs in the flow pick different weight vectors to generate their trajectories, in conjunction with the cooperation factor considering social behavior in the decision module, bringing diversity to the whole vehicle flow.

\section{Experiment}
\label{sec:experiment}

\begin{figure}[tbp]
  \centering
  \includegraphics[width=\linewidth]{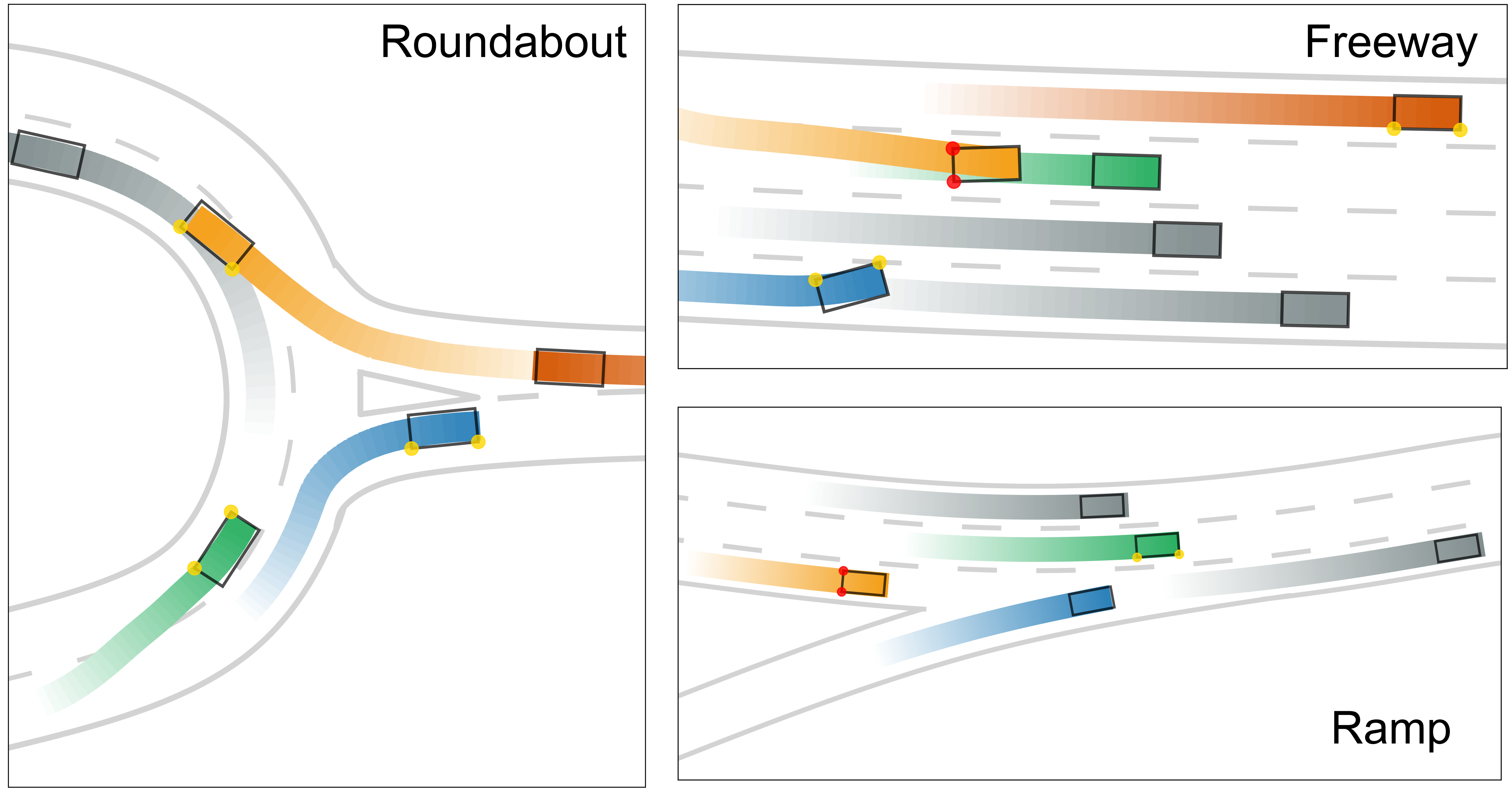}
  \caption{Vehicle trajectories in the different scenarios}
  \label{fig:scenario}
  \Description{.}
\end{figure}

We utilize real-world driving records from the CitySim dataset \cite{ou2022} to generate the experimental environment, which consisted of three different scenarios: Freeway, Ramp, and Roundabout, as shown in Figure~\ref{fig:scenario}.
The Freeway scenario is the simplest one, containing a four-lane expressway. The ramp scenario includes a three-lane main road with an on-ramp merging into it.
The roundabout  scenario measures the performance of vehicle flows at an unsignalled two-lane roundabout entrance.
We take the vehicle information from the dataset as our experiment's initial state of the vehicle flow.
The vehicles about to make a lane change (including entering and exiting the ramp) and the vehicles directly interacting with them  are set as AVs controlled by the centralized planner. In contrast, other unaffected vehicles are set as uncontrolled HVs. The vehicles' lane change intentions are fed into the framework as input.
The proposed framework is implemented entirely in Python for fast prototyping.

% 列出所有三种不同行为的参数
The vehicle size in the flow is set to 5m $\times$ 2m.
The constant $C_p=1/\sqrt{2}$ in Equation~\ref{eq:UCT}.
We set reaction time $\tau=0.5$s and minimum time headway $MTH=3$s in Equation~\ref{eq:safe distance}.
The fixed acceleration $a_{acc}$ and deceleration $a_{dec}$ in available actions of Table~\ref{tab:actions} are set to 0.6m/s$^2$.
The time step is set to 1.5s in the decision module and 0.1s in the planning module.
To generate vehicle flows with diversity, three different types of driving behavior are defined: aggressive, normal and conservative.
The aggressive vehicle takes less account of the surrounding vehicles when making decisions and thus has a smaller cooperation factor $\gamma$, along with a higher tolerance to both obstacle avoidance and acceleration changes. The opposite is true for the conservative vehicles.
The specific style-related parameters are defined in Table~\ref{tab:types}.
\begin{table}[tbp]
  \caption{Three types of driving behaviors}
  \label{tab:types}
  \begin{tabular}{cccc}
    \toprule
    Type         & Aggressive & Normal & Conservative \\ \midrule
    $\gamma$     & 0.1        & 0.5    & 0.9          \\
    $w_{i,obs}$  & 2.0        & 4.0    & 6.0          \\
    $w_{i,out}$  & 6.5        & 5.0    & 4.0          \\
    $w_{i,jerk}$ & 0.7        & 1.0    & 1.2          \\ \bottomrule
  \end{tabular}
\end{table}
%其他设置相同

\begin{table*}[tbp]
  \caption{Comparative experiments in the Freeway scenario}
  \label{tab:compare}
  \begin{tabular}{l|ccc|ccc|ccc}
    \toprule
                                               & \multicolumn{3}{c|}{2 Controlled AVs}                    & \multicolumn{3}{c|}{3 Controlled AVs} & \multicolumn{3}{c}{4 Controlled AVs}                                                                                \\
                                               & \multicolumn{1}{c}{\rotatebox{0}{\makecell[c]{Sequential                                                                                                                                                               \\MCTS}}} & {\rotatebox{0}{\makecell[c]{Ours w/o\\Planning}}} & \multicolumn{1}{l|}{\rotatebox{0}{Ours}} & \multicolumn{1}{c}{\rotatebox{0}{\makecell[c]{Sequential\\MCTS}}} & {\rotatebox{0}{\makecell[c]{Ours w/o\\Planning}}} & \multicolumn{1}{l|}{\rotatebox{0}{Ours}}  & \multicolumn{1}{c}{\rotatebox{0}{\makecell[c]{Sequential\\MCTS}}} & {\rotatebox{0}{\makecell[c]{Ours w/o\\Planning}}} & \multicolumn{1}{l}{\rotatebox{0}{Ours}} \\ \midrule
    Sucess Rate~$\uparrow$                     & 100\%                                                    & 100\%                                 & \textbf{100\%}                       & 100\%  & 100\%  & \textbf{100\%}  & 70\%          & 90\%   & \textbf{100\%}  \\
    Avg Expand Nodes $\downarrow$              & 1304.1                                                   & 1196.6                                & \textbf{1148.7}                      & 1387.6 & 1127.8 & \textbf{1037.6} & 1706.6        & 1785.0 & \textbf{1527.3} \\
    Avg Finish Time (s) $\downarrow$           & \textbf{3.57}                                            & 5.75                                  & 5.68                                 & 3.57   & 4.95   & \textbf{3.50}   & \textbf{4.58} & 9.29   & 8.93            \\
    {\makecell[l]{Min Distance (m)}$\uparrow$} & 4.34                                                     & 8.41                                  & \textbf{9.13}                        & 4.74   & 7.02   & \textbf{10.78}  & 5.02          & 7.54   & \textbf{10.94}  \\ \bottomrule
  \end{tabular}
\end{table*}

Our proposed framework generates promising planning solutions within all three scenarios.
The generated traffic trajectories are shown in Figure~\ref{fig:scenario}, where the controlled AVs in the flow is denoted by colored trajectories and uncontrolled HVs by grey trajectories.
% Brake lights and turn signals are also displayed in the figure.
The interpretable actions generated by the decision module are shown in the figure through the brake lights and turn signals.

We further compare and analyze the performance of our proposed framework and the diversity of the solutions it generates in the following sections.

%decision 模块会对每个时间步都进行一次MCTS搜索，而不是只从根节点进行一次
\subsection{Framework Performance}

In this subsection, we test the performance of the proposed framework.
To evaluate the effectiveness of our proposed method, we first implemented a sequential MCTS approach as a comparison, which is used in~\cite{Li2022} (marked as \textit{Sequential MCTS}). This method assigns a right-of-way priority to each vehicle in the flow and makes decisions for vehicles in order of priority at each decision step. In this experiment, vehicles with greater longitudinal distance possess a higher decision priority.
As the previous study did not consider uncontrolled HVs in the scenario, we employ a car-following model~\cite{treiber2000congested} to predict all HVs' trajectories and feed them into the sequential MCTS as dynamic obstacles.
We also include a decision-only framework (marked as \textit{Ours w/o Planning}) in the comparison experiments to demonstrate the necessity and effectiveness of the planning module in our proposed framework.

According to~\cite{treiber2000congested}, the longitudinal acceleration in sequential MCTS is chosen to be [$-1,0,1$]m/s$^2$, and the lateral velocity is chosen to be [$-1.2,0,1.2$]m/s.
We set the decision time step for sequential MCTS and decision-only framework to 1.0s.
The planning horizon for all three methods is 10s and the maximum iteration number of MCTS is 3000.

The experiments are conducted in the four-lane freeway scenario with the number of controllable AVs set to 2, 3, and 4.
Each method is repeated ten times under the same experimental setup.
We measure each method's success rate, the average number of expansion nodes in MCTS, the average time to finish the vehicle's intention and the minimum inter-vehicle distance in the generated solutions.
The results are shown in Table~\ref{tab:compare}.

All three methods achieve a 100\% success rate in the freeway scenarios with two and three AVs.
However, when the number of controlled AVs reaches four, sequential MCTS can only attain a success rate of 70\%. The decision-only approach has a 90\% success rate, while our framework still manages to pass all tests. As the sequential MCTS defines the priority for vehicles in the flow, the higher-priority vehicles do not yield to the lower-priority ones when making decisions. This results in the high-priority vehicles taking locally optimal decisions that make it impossible for the subsequent vehicles to complete their intentions.

The number of extended nodes in the search tree measures the search efficiency of each method. Our proposed framework has fewer available actions for each vehicle and adopts various pruning strategies as we mentioned in Section~\ref{prune strategy}.  Experimental results show that our approach expands fewer nodes than the comparison methods, which means that our framework searches for the optimal solution faster without wasting actions on infeasible actions.

\begin{figure*}[tbp]
  \centering
  \includegraphics[width=0.899\linewidth]{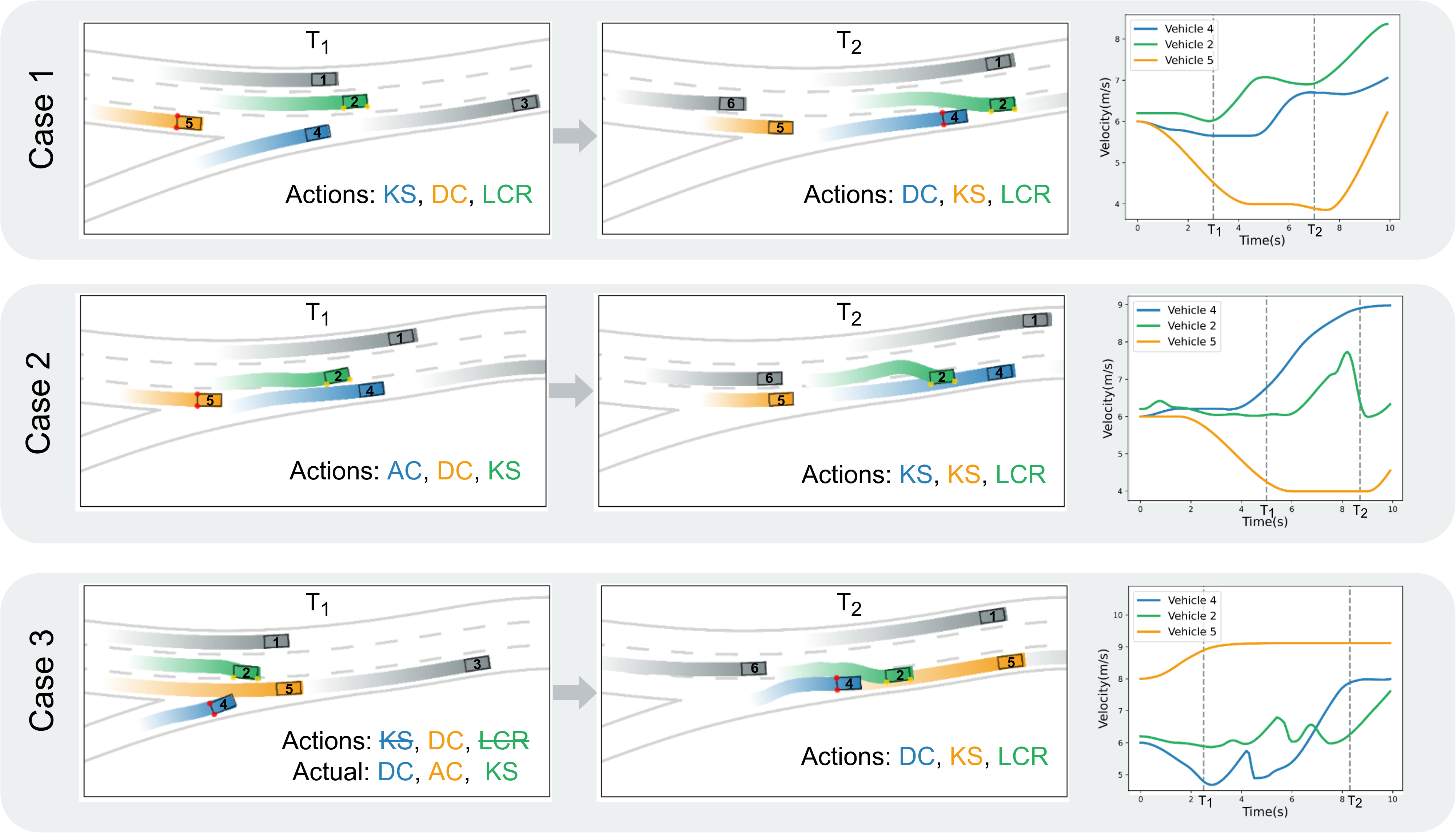}
  \caption{Three different cases in the closed-loop simulation}
  \label{fig:case study}
  \Description{This diagram demonstrates the pipeline of a vehicle trajectory planner.}
\end{figure*}
\begin{figure}[tbp]
  \centering
  \includegraphics[width=0.7\linewidth]{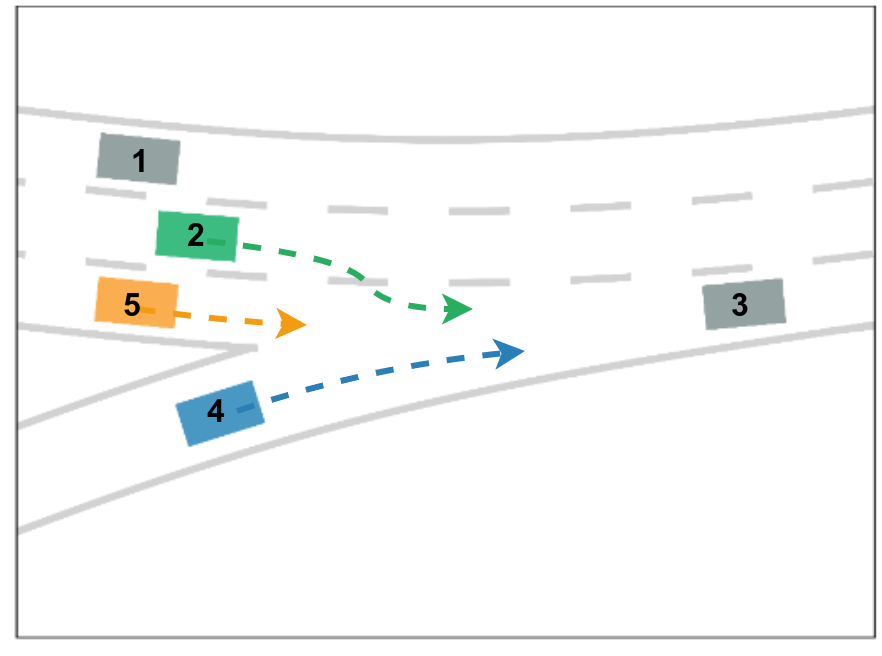}
  \caption{Vehicles' initial states in closed-loop simulation}
  \label{fig:close-loop}
  \Description{This diagram demonstrates the pipeline of a vehicle trajectory planner.}
\end{figure}

In terms of the quality of the solutions, although the average time for vehicles to complete their intentions is shorter in the solutions generated by the sequential MCTS, the minimum inter-vehicle distances of these solutions are shorter and hence carry more risk.
The shortest distance between vehicles in the solution planned by sequential MCTS is generally less than 5 meters (the length of the vehicle), which can easily lead to rear-end collisions.
Our decision module, in contrast, achieves a minimum distance of seven to eight meters thanks to introducing the shortest safe distance $D_s$.
The planning module refines the trajectory further, bringing the whole approach to a minimum distance of around 10 meters, i.e., twice the vehicle length.
Besides, the whole framework has a shorter average intention finishing time compared to the decision-only approach, even outperforming the sequential MCTS in the case of 3 vehicles.

\subsection{Closed-loop Simulation}

We conducted several closed-loop simulations in the ramp scenario to investigate the diversity of the vehicle trajectories regarding social behavior and driving habits. We also demonstrate the proposed framework's responsiveness when facing human drivers' unanticipated actions.

The closed-loop simulation runs at 10Hz. Our decision-making module has a horizon of 10.5 seconds, with 1.5 seconds for each action. The horizon of the planning module is 3 seconds with a time step of 0.1 seconds, and the replanning cycles are 0.5 seconds.
Our test scenario is shown in Figure~\ref{fig:close-loop}.
Vehicles with the number 2, 4, and 5 are AVs controlled by our framework, while vehicles with the number 1, 3, and 6 are HVs in the environment.
Vehicles 2 and 5 are driving side by side in the adjacent lanes of the main road, while vehicle 4 is on the ramp about to merge onto the main road.
In addition, vehicle 2 intends to change its lane to the right, creating a complex situation where three cars in the flow compete for the right-of-way of the rightmost lane simultaneously.
Figure~\ref{fig:case study} shows three different cases arising from this same initial state. The velocity profiles of controlled AVs in each case are shown on the right of the figure.
% The interpretable actions produced by the decision module are also displayed on the diagram.

In Case 1, vehicle 2 is assigned an aggressive driving behavior, vehicle 4 adopts a conservative driving behavior and vehicle 5 is assigned the normal behavior. Their style-related parameters are set according to Table~\ref{tab:types}.
The closed-loop experimental results show that due to its backward initial position, vehicle 5 slows down to give way to vehicle 4, who maintains its initial velocity and merges onto the main road.
At the moment $T_1$, as vehicle 2 has a more aggressive driving behavior, it is unlikely to yield to vehicle 4.
Vehicle 2 continues to drive in the left front of vehicle 4 and indicates its intention to change lanes. Finally, vehicle 2 slows down to make space for vehicle 4 to complete its lane change at $T_2$.

We change the driving behavior of the Avs in Case 2. The driving behavior of vehicle 4 is altered to aggressive and vehicle 2 is adjusted to conservative. The driving behavior of vehicle 5 remains the same.
Vehicle 5 still slows down to let vehicle 4 merge into the main road at $T_1$, but this time vehicle 4 does not yield to vehicle 2's lane change and chooses to accelerate through it.
Vehicle 2, aware of vehicle 4's acceleration, chooses to slow down and completes the lane change action after vehicle 2, which is quite different from Case 1.

In Case 3 we explore what happens when an uncontrollable HV in the environment does not cooperate with AV's behavior.
We change vehicle 5 from AV to HV that is not controlled by our planner and set a larger initial speed for it.
Instead of slowing down to give way to vehicle 2, as our framework expected, it accelerates into the ramp entrance, creating a quite dangerous situation.
Thanks to the high-frequency replanning in our planning module, vehicle 2 and vehicle 4 abort their original actions from the decision module when they realize that vehicle 5 is accelerating in order to avoid the potential collision.
Vehicle 4 performs a deceleration and vehicle 2 keeps its original lane.
Once vehicle 5 has passed, vehicle 2 and 4 then complete their lane-changing and merging behavior safely.
This case demonstrates the importance of integrating the planning module into the framework, which can quickly respond to changes in the road and resolve potential collisions between vehicles, safeguarding the entire vehicle flow.

\section{Conclusions}
\label{sec:conclusion}

This paper presents a novel hierarchical multi-vehicle decision-making and planning framework that produces trajectories with diversity, interactivity, and interpretability in mixed AV and HV scenarios.
By introducing the vehicle's cooperation factor in the decision module and the trajectory weight set in the planning module,  the trajectories generated by our framework are diverse at both the social interaction and individual habit levels.
Our framework makes decisions jointly for all vehicles in the flow and ensures a real-time response to the dynamic environment through a high-frequency re-planning method.
Also, the decision module generates interpretable action sequences, which are leveraged to inform passengers of the current situation and explicitly communicate self-intents to the surrounding HVs.

In the further work, our framework can be used to generate background traffic around the ego car in the autonomous driving simulator. It can also be utilized to perform data augmentation of the existing traffic datasets with different driving styles.
In addition, our framework is generic in nature, meaning each module's internal algorithms can be replaced, provided their respective inputs and outputs are satisfied.

%%%%%%%%%%%%%%%%%%%%%%%%%%%%%%%%%%%%%%%%%%%%%%%%%%%%%%%%%%%%%%%%%%%%%%%%

%%% The acknowledgments section is defined using the "acks" environment
%%% (rather than an unnumbered section). The use of this environment 
%%% ensures the proper identification of the section in the article 
%%% metadata as well as the consistent spelling of the heading.

% \begin{acks}  
%   This research was supported by the Science and Technology Commission of Shanghai Municipality (Grant Nos. 22YF1461400 and 22DZ1100102).
% \end{acks}

%%%%%%%%%%%%%%%%%%%%%%%%%%%%%%%%%%%%%%%%%%%%%%%%%%%%%%%%%%%%%%%%%%%%%%%%

%%% The next two lines define, first, the bibliography style to be 
%%% applied, and, second, the bibliography file to be used.

\balance
\bibliographystyle{ACM-Reference-Format}
\bibliography{reference}

%%%%%%%%%%%%%%%%%%%%%%%%%%%%%%%%%%%%%%%%%%%%%%%%%%%%%%%%%%%%%%%%%%%%%%%%

\end{document}